# A Data Bootstrapping Recipe for Low-Resource Multilingual Relation Classification

**Arijit Nag**
IIT Kharagpur
arijitnag@iitkgp.ac.in

**Bidisha Samanta**
IIT Kharagpur
bidisha@iitkgp.ac.in

**Animesh Mukherjee**
IIT Kharagpur
animeshm@cse.iitkgp.ac.in

**Niloy Ganguly**
IIT Kharagpur, Leibniz University Hannover
niloy@cse.iitkgp.ac.in

**Soumen Chakrabarti**
IIT Bombay
soumen@cse.iitb.ac.in

## Abstract

Relation classification (sometimes called 'extraction') requires trustworthy datasets for fine-tuning large language models, as well as for evaluation. Data collection is challenging for Indian languages, because they are syntactically and morphologically diverse, as well as different from resource-rich languages like English. Despite recent interest in deep generative models for Indian languages, relation classification is still not well-served by public data sets. In response, we present IndoRE, a dataset with 21K entity- and relation-tagged gold sentences in three Indian languages, plus English. We start with a multilingual BERT (mBERT) based system that captures entity span positions and type information, and provides competitive monolingual relation classification. Using this system, we explore and compare transfer mechanisms between languages. In particular, we study the accuracy-efficiency tradeoff between expensive gold instances vs. translated and aligned 'silver' instances. We release the dataset for future research.[1]

## 1 Introduction

Relation classification (sometimes called relation 'extraction' or RE) is the task of identifying a semantic relation (from a catalog of canonical relations) that holds between two nominal entities in text. Related to semantic role labeling, it is a critical capability for question answering and knowledge graph (KG) augmentation. The key challenge in RE is the diversity of textual expressions of canonical relations. WikiData, among the largest public KGs, has a hundred million entities with aliases but only thousands of well-instantiated canonical relations, with relatively sparse textual descriptions of relations, particularly in low-resource languages.

Low-resource languages, including many Indian languages, neither have well-tuned standard NLP pipelines, nor unsupervised corpora as large as resource-rich languages. Not only are they often morphologically distant from resource-rich language families, but even their syntax for relation expression differs. E.g., while subject-predicate-object (infix) is more common in English, subject-object-predicate (postfix) is more common in Indic languages: "Ravi went to Delhi" (En) vs. रवि दिल्ली गया था/"Ravi Delhi went-did" (Hi). Word order is more relaxed: दिल्ली गया था रवि/"Delhi went-did Ravi" is acceptable; only the entity types help orient the predicate. Verb particles are more common than in English: "Talk to him" (En) becomes उससे बात करो/ "him-to talk do" (Hi) or তার সাথে কথা বলুন/"him with speech talk" (Bn). Transliteration and code switching are rampant: "I called him yesterday" (En) translates to " मैंने कल उसे कॉल किया था"/"I yesterday him call did-past", where *call* is transliterated.

Against this backdrop of diverse textual expression of relational facts, collecting high-quality labeled data for Indian language RE is a major challenge, particularly given data-hungry state-of-the-art neural RE systems. Responding to these pressures, we contribute a multilingual RE testbed for low-resource Indian languages. Specifically, we build and contribute **IndoRE**, a diverse and rich set of entity- and relation-annotated sentences in three Indian languages — Bengali (Bn), Hindi (Hi) and Telugu (Te). In addition, we provide labeled English (En) RE instances. Our motive is to study and catalog a diversity of protocols for transferring RE capability across languages, chiefly in the resource-rich to resource-poor direction, starting from the "each language for itself" (ELFI) operating point. Owing to distributional fidelity, we generally expect that, with 'sufficient' gold training data, ELFI will show strong performance; the catch being the expense of collecting gold in-stances in resource-poor languages.

---
[1] https://github.com/NLPatCNERG/IndoRE

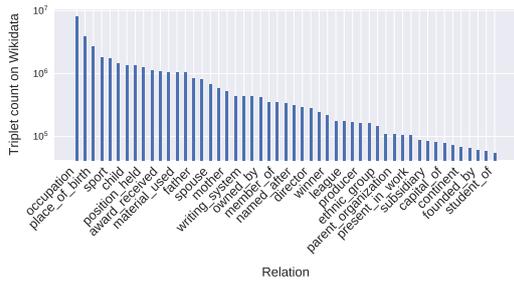

Figure 1: Number of KG triples involving the selected relations.

To investigate the extent to which various modes of transfer can compensate for gold instances, we start by designing a competitive base RE system built on mBERT.[2] By sharing a multilingual wordpiece vocabulary and training on a vast multilingual corpus, mBERT already lends some support to go beyond ELFI. Beyond this baseline, we explore ways to train models for RE in a target language by 'borrowing' training instances from a given source language, translating and aligning, and preparing 'silver' instances to train an RE system for the target language. Such *model* transfer (Kozhevnikov and Titov, 2014) may be attempted in a zero-shot mode, where no gold-labeled instances are available in the target language, or in a few-shot mode where a few such instances are prepared at some cost. We also explore an *instance* transfer setting, in which we train the RE module using only English instances, and translate target language test instances to English as well.

| Relation ($r$): | spouse |
|---|---|
| Sentence ($s$): | विराट कोहली और अनुष्का शर्मा ने 2017 में इटली में शादी कर ली थी |
| English: | Virat Kohli and Anushka Sharma got married in Italy in 2017. |
| Entity pair: | विराट कोहली , अनुष्का शर्मा |
| Lexical distance: | 1 |
| Relation ($r$): | award_received |
| Sentence ($s$): | बृहत्–पिंगला गुजराती भाषा के विख्यात साहित्यकार रामनारायण पाठक द्वारा रचित एक छंद शास्त्र है जिसके लिये उन्हें सन् 1956 में गुजराती भाषा के लिए मरणो–परांत साहित्य अकादमी पुरस्कार से सम्मानित किया गया . |
| English: | Brihat-Pingala is a verses scripture composed by noted Gujarati litterateur Ramnarayan Pathak for which he was posthumously awarded the Sahitya Akademi Award for Gujarati language in 1956. |
| Entity pair: | बृहत्–पिंगल , साहित्य अकादमी पुरस्कार |
| Lexical distance: | 24 |

Table 1: Sample 'simple' and 'complex' relations.

---

[2] https://github.com/google-research/bert/blob/master/multilingual.md

The presence of multiple languages, with diverse levels of similarity between them, brings a new dimension to our study. We find that 0-shot model transfer works better between Indian languages than from English to an Indian language, but is still behind ELFI by 11.27% (Bn), 11.06% (Hi) and 13.43% (Te) macro F1. Additional fine-tuning with 10-shot target gold instances reduces the gap to 3.76% (Bn), 2.8% (Hi) and 4.23% (Te). Effectiveness of model transfer also shows interesting and intuitive patterns and variations across language pairs in our transfer framework, **TransRel**. We observe that translating test instances to English helps a model trained on English instances perform well; however, the gain diminishes as we add a few training samples translated to English from the target language.

## 2 IndoRE data collection process

We focus on three Indian languages: Hindi (Hi), Bengali (Bn) and Telugu (Te). Our intention was to pick two languages (Hi and Bn) that are more similar to each other than to the third language, Te. However, each of these languages has certain distinctive features not found in the others. As an example of a resource-rich language, we include English (En).

In this section, we first discuss the limitations of existing multilingual RE data sets. Then we describe the data collection process, starting with relation selection, and followed by evidence sentence collection, data processing and automatic annotation. Then we discuss annotation evaluation and cleanup using human annotators.

### 2.1 Available datasets and limitations

Several groups (Hendrickx et al., 2010; Han et al., 2018; Riedel et al., 2010; Zhang et al., 2017) have released distant-supervised or human-annotated RE datasets in English. As we shall argue, translating them to respective Indian languages may not generate a good representative collection. The human-curated supervised datasets **SemEval'10** (Hendrickx et al., 2010) and **FewRel** (Han et al., 2018) contained ∼10k and 56k sentences respectively over nine and 80 relations. The relations for FewRel are from Wikipedia. As the relations mainly focus on an English corpus, most of them do not have sufficiently many representative candidates in Indian languages as they are not relevant to the Indian context. E.g., FewRel has a relation taxon_rank with sample passage "…more formally brought together the three families, Agapanthaceae, Alliaceae, Amaryllidaceae, under the single Asparagalean monophyletic family, now renamed Amaryllidaceae from Alliaceae, reversing the Dahlgrenian process of family splitting…". This relation has no candidate in any Indian language.

The distant-supervised dataset **TACRED** (Zhang et al., 2017) released relations which are not from existing knowledge-bases like Wikipedia or DBpedia. TACRED was created based on query entities and annotated system responses in the yearly TAC KBP evaluations. Therefore, this data is very distinct and specific to the TAC KBP corpus, resulting in certain cultural gaps. For example "Billy Mays, … pop culture icon, died at home in Tampa." is evidence for the relation city_of_death between entities "Billy Mays" and "Tampa". In Bengali, মারা যান means 'died' for commonors, but for celebrities, শেষ নিঃশ্বাস ত্যাগ করেন (last breath release did) is more common. These cultural variations cannot be captured by translating sentences very specific to En.

The **NYT** data set (Riedel et al., 2010) includes 42 relations from Freebase, but entity types are restricted to only businesses, people and locations, which makes it restrictive and less diverse. We provide a comparison among the datasets in Table 2.

| Dataset | Language | KB | #Relation | #Sentence |
|---|---|---|---|---|
| SemEval'10 | En | - | 9 | 10,717 |
| FewRel | En | Wikidata | 80 | 56,000 |
| TACRED | En | TAC KBP | 42 | 119,474 |
| NYT | En | Freebase | 53 | 126,184 |
| **IndoRE** | En,Bn,Hi,Te | Wikidata | 51 | 32,610 |

Table 2: Comparison of different datasets.

### 2.2 Relation selection

We select 51 relations from WikiData using stratified sampling, proportional to the number of available (subject, relation, object) triples per relation, including all languages present in WikiData. Figure 1 shows triplet counts in decreasing order for various relations. Some relations are (linguistically) easy to express, e.g., spouse, whereas other relations are complex and need longer sentences to express, e.g., award_received; see Table 1. One rough indicator of the complexity of a relation is the number of tokens (lexical distance) between the two related entities. For various relations, the average lexical distance in different languages is shown in Appendix A, Figure 3.

### 2.3 Evidence sentence collection

After selecting the relations, we collect potential evidence sentences for each relation in each language. As there is no direct way to obtain large numbers of gold evidence sentences, we apply distant supervision (Mintz et al., 2009).

We sample a set of entity pairs for each relation in each language using stratified sampling proportional to the sentence count per entity pair. We collect a potentially noisy set of evidences per entity tuple, i.e., per $(e_1, e_2)$, by collecting a maximum of 1000 sentences, based on keyword search.

The entire English, Bengali, Hindi and Telugu Wikipedia dumps[3] are processed using Pyserini[4], an open-source information retrieval toolkit. We query its index with entity pair $(e_1, e_2)$ separated by space, and collect the top-1000 results. The intuition of such distant supervision is that if two entities participate in a relation, any sentence which mentions these two entities has some chance of expressing it. To enforce data diversity, we further collect sentences from mixed sources, by querying the Google Web search engine API with $(e_1, e_2)$ and selecting the top five response URLs,[5] which are fetched. From the HTML, we extract sentences where both the entities are present.

### 2.4 Data filtering and annotation

Distant supervised methods can yield noisy sentences, i.e., even though they contain the desired entity pair $(e_1, e_2)$, they may not provide evidence of the relation $r$. We filter out some non-evidence sentences as follows. We generate an embedding ($emb_s$) of each collected sentence — after removing the two entity mention spans — as an average word embedding using mBERT. Similarly, we obtain relation embedding ($emb_r$) as an mBERT embedding of the relation by taking average over all the individual contextual embedding of the description and alias of a particular relation from WikiData. Given a sentence $s$ and corresponding desired relation $r$, we retain sentences for which $\cos(emb_s, emb_r) \geq \tau$, where $\tau$ is a threshold which is empirically determined for each language (typically, 0.3 gave the best results).

Next we mark the entity spans in each sentence by two special marker pairs $[E_1][/E_1]$ and $[E_2][/E_2]$ as it is useful to learn a generalized embedding for entities which can help in relation extraction (Soares et al., 2019). For English sentences, we obtain entity types using the spaCy NER tagger.[6] For Indian languages we translate the entity into English and obtain the entity type with the same tool.

---
[3] https://dumps.wikimedia.org/
[4] https://github.com/castorini/pyserini/
[5] https://pypi.org/project/google/

| Language | #Sentence | #Distinct entity pair |
|---|---|---|
| En | 8486 | 7232 |
| Bn | 8291 | 5321 |
| Hi | 6979 | 4399 |
| Te | 8854 | 4041 |
| Total | 32610 | 20993 |

Table 3: Salient statistics of IndoRE.

## 2.5 Annotation evaluation

After applying the preprocessing steps described above, we clean the instances by employing human annotators. We assign instances to annotators who are fluent in the language of the instance. We provide the canonical relation name ($r$), a sentence $s$ marked with entity pair spans, entity pair ($e_1, e_2$) and entity types ($et_1, et_2$). A sample is shown (with En translation added).

| Relation ($r$): | spouse |
|---|---|
| Sentence ($s$): | $[E_1]$ मिशेल ओबामा $[/E_1]$ अमेरिका की पूर्व राष्ट्रपति $[E_2]$ बराक ओबामा $[/E_2]$ की पत्नी हैं , एवं अमेरिका की प्रथम महिला रह चुकी हैं |
| English translation: | Michelle Obama is the former first lady and wife of former American president Barack Obama |
| Entity pair: | मिशेल ओबामा, बराक ओबामा |
| Entity type pair: | Person, Person |

We also provide a list of entity types and necessary context by providing a relevant link to the page where the relation definition and other related information are explained. The annotators are asked to perform the following tasks.

(a) Discard the sentence if it does not entail relation $r$ between $e_1$ and $e_2$.
(b) Correct the entity spans $[E_1] \cdots [/E_1]$ and $[E_2] \cdots [/E_2]$ if necessary.
(c) Correct the entity types $et_1$ and $et_2$ consulting the list of entity types, if needed.

We first ran a pilot study with 100 sentences for each language and found that the inter-annotator agreement was more than 85%. Thereafter, we processed the rest of the data with one annotator per instance. Statistics of the final cleaned data are provided in Table 3.

## 2.6 Silver instance generation

To increase coverage of examples for the relations located at the tail of the distribution as reported in Figure 1, we also generate (semi-) synthetic or 'silver' data that is used for model transfer from one language to another, commonly from a resource-rich to low-resource language. We illustrate the process using En and Hi.

We translate gold-labeled En instances into Indian languages (say, Hi) using the Google translate API.[7] As there is no commonly available NER tagger in Indian languages, we use English entities to mark the entity span in Indian languages. We translate the entities of the source English sentence into Hi and mark the entity spans in the target synthetic sentence using a simple heuristic described in Appendix B. Finally, we borrow the entity types from the source sentence. The following example walks through this process in detail. As we do not rectify the entity mapping using human annotators, we term this set as silver data.

| English | Virat Kohli and Anushka Sharma got married in Italy in 2017. |
|---|---|
| Entity (Type) | Virat Kohli (Person), Anushka Sharma (Person) |
| Translated Hindi | विराट कोहली और अनुष्का शर्मा ने 2017 में इटली में शादी कर ली थी |
| Entity mapping | विराट कोहली (Person), अनुष्का शर्मा (Person) |

## 3 Base RE module

We build a base RE module which will be used for RE in Indian languages with various cross lingual transfer schemes. As depicted in Figure 2, our architecture is built on multilingual BERT (mBERT), a masked language model.[8]

For effective relation extraction in a multilingual setting, at the input layer, we provide entity tagged inputs; the summarization layer collects important information about the sentence and entities; finally, the relation extraction layer predicts the relation using entity type information along with other contextual information. We try different variations of the architecture and describe the best preforming variation here. We will describe some important details of preparing the input, the summarization, and the output RE layers.

### 3.1 Input layer

We insert special marker tokens in the input sentence ($s$) for effective presentation to mBERT. We mark entity spans by special tokens $E_1$ and $E_2$ as

---
[6] https://spacy.io/
[7] https://cloud.google.com/translate
[8] https://github.com/google-research/bert/blob/master/multilingual.md

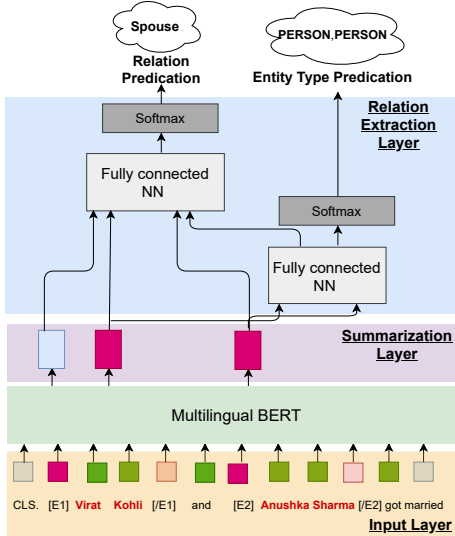

Figure 2: Base RE system, consisting of an input layer, multilingual LM (mBERT), contextual embedding summarization layer, and relation extraction layer.

follows:

$$s = ([CLS], w_1, \ldots, [E_1], e_1, [/E_1],$$
$$w_i, \ldots, [E_2], e_2, [/E_2], \ldots, [SEP]) \quad (1)$$

where $e_1$ and $e_2$ represents mention spans of the two entities of interest and $w_i$ represents the $i^{\text{th}}$ token/word. We provide details of annotating the input sequence with special markers for entity types in Appendix C.

### 3.2 Summarization layer

mBERT outputs contextual embeddings ($h$) for each original token, as well as the special tokens introduced in the input sentence $s$. We pool the contextual information from key positions (Ni et al., 2020) as follows.

Along with CLS which consists of overall sentence information, we also consider the entity-specific information from the entity span tokens and all the tokens between them.

$$h_s = [h_{[CLS]}, h_{E_1}, h_{E_2}] \quad (2)$$
$$h_{e_1}^d = \operatorname*{avg}_{i \in [E_1, \ldots /E_1]} h_i^d, \quad \forall d \in [1, D] \quad (3)$$
$$h_{e_2}^d = \operatorname*{avg}_{i \in [E_2, \ldots /E_2]} h_i^d, \quad \forall d \in [1, D] \quad (4)$$

where $D$ is the embedding dimension.

We compare two potential summarizing mechanism (a) **CLS+ES** which consists of infomation in $h_s$ and (b) **CLS+EN** which consists of $h_{[CLS]}$ and $h_{e_1}, h_{e_2}$. Other architectural variations are discussed in Appendix C.

### 3.3 Relation extraction layer

Embeddings from the summarization layer can be used in a couple of ways to obtain relation labels.

**RE (Relation Extraction):** We use **CLS + ES** or **CLS + EN** to classify the relations. We do not fine-tune our model for entity classification.

**ME (multitasking ensemble):** We ensemble the best performance of the following two mechanism per relation.

*Without parameter sharing:* We use $h_{e_1}$ and $h_{e_2}$ to classify entity types and $h_s$ to separately classify relation $r$.

*With parameter sharing:* We obtain entity type embedding $h_{et_1}$ and $h_{et_2}$ by transforming $h_{e_1}$ and $h_{e_2}$ respectively and use them to predict entity types. We further use them along with $h_s$ to predict relation $r$.

## 4 Interlingual transfer mechanisms

By using mBERT, the base RE system already exploits generalized language agnostic information (Pires et al., 2019). In addition, our contextual summaries and type encodings also help map entities and relations in different languages to a common space. Over and above these, we explore transfer through machine translation.

**ELFI (Each language for itself):** The base RE system is trained using gold instances from the target language, and tested on other instances in the same language. At a high labeling cost, this gives an idea of the maximum achievable RE quality.

**LMx (Transfer via mBERT):** The base RE system is trained with source language instances and zero target language instances. Then it is tested with target language instances. This is the zero-shot setting **LMx0**. The whole burden of transfer is on mBERT's language model(LM) in this case. In a related few-shot setting, we allow 1, 5, and 10 sentences per relation from the available gold target instances to fine-tune the RE model. These are called **LMx1**, **LMx5** and **LMx10**. This lets us calibrate the value of gold target language instances vis-a-vis many training instances in the (resource-rich) source language.

**MTx (Model Transfer):** Instead of forcing mBERT to do all the heavy lifting, we help out RE at training time by translating (plentiful) gold source language instances into silver target language instances (as described in Section 2.6) to train RE in the target language. This form of translation is widely used for model transfer, see Kozhevnikov and Titov (2014), *inter alia*. Here, too, we can throw in zero or a few target gold instances, leading to variations we call **MTx0**, **MTx1**, **MTx5** and **MTx10**.

**Ix (Instance Transfer):** Here we go the other way, from the low-resource target to the resource-rich source language, at test time. At training time, we train the base RE system only on English instances (some of which could be translated from target training instances). Each test instance in the target language is translated and aligned (see Appendix B) to English, after which it is processed by the English-trained base RE system. In the zero shot setting **Ix0**, no gold target instances are used. In a few-shot setting **Ix10**, we translate 10 gold target sentences per relation to English as additional training data for fine tuning. Our goal is to study the performance gain from **Ix0** to **Ix10**.

## 5 Interlingual transfer performance

For each language, we split the gold data for each relation into training (80%) and testing (20%) folds. We keep the test fold fixed. We ensure *zero overlap* in entity pairs for every relation $r$ between the folds, and report on 3-way cross validation. We compare the **RE** and **ME** task variants and collect macro F1 scores (shown here) and 0/1 accuracy (given in Appendix D). They show the same trend across methods.

| Summarization | Tasks | Bn | Hi | Te |
|---|---|---|---|---|
| CLS+ES | RE | 92.20 | 94.21 | 72.42 |
| | ME | 92.33 | 94.29 | 72.23 |
| CLS+EN | RE | 92.09 | 93.95 | 73.71 |
| | ME | 91.81 | 94.60 | 75.43 |

Table 4: RE macro F1 for **ELFI**.

### 5.1 Performance of ELFI

Table 4 reports macro F1 for **ELFI**. We observe that both the summarization layer variations yield comparable results. Therefore, for the rest of the experiments, we will report **CLS+ES**. Multitasking ensembles (ME) provide better accuracy for all the languages. Macro F1 is notably lower for Telugu. After human filtering, several relations fell below 20 instances, resulting in insufficient training.

| ↓Source | Tasks | Bn | Hi | Te |
|---|---|---|---|---|
| ELFI (best) | | 92.33 | 94.60 | 75.43 |
| En | RE | 63.37 | 67.65 | 42.53 |
| | ME | 66.84 | 69.37 | 45.03 |
| Bn | RE | - | 81.45 | 57.01 |
| | ME | - | 80.39 | 58.96 |
| Hi | RE | 76.23 | - | 56.78 |
| | ME | 75.09 | - | 55.14 |
| Te | RE | 60.28 | 60.48 | - |
| | ME | 62.12 | 66.37 | - |
| ALL | RE | 79.68 | 83.45 | 61.58 |
| | ME | 81.06 (-11.27) | 83.54 (-11.06) | 62.00 (-13.43) |

Table 5: Macro F1 of LMx0. Green represents best and Red represents the worst performance compared to ELFI. Yellow represents the best performance for a single source language.

### 5.2 Performance of LMx0

Table 5 shows LM-based transfer results. Of considerable interest is the pairing of (source, target) languages that result in best performance.

- Source Hi helps to achieve the best macro F1 score for target Bn, compared to other languages. Conversely, Bn is the best source for Hi. This may be explained by their membership in the Indo-European language genus.
- In contrast, Te, belonging to the Dravidian genus, is indifferent to Hi vs. Bn.
- Source En performs poorly for all three Indian languages. So it seems better for Indian languages to assist each other for RE, compared to importing foreign help.
- That being said, the best target performance is uniformly achieved by training on *all sources except the target*, which contributes diversity from multiple language families and builds a robust RE model.
- However, this best performance is still substantially behind ELFI performance (last row) for all languages.

### 5.3 Performance of LMx{1, 5, 10}

Table 6 shows the effect of allowing a small percentage of available gold instances in the target language for training. Understandably, revealing more instances is better; LMx10 gives the best target performance. The deficit from ELFI reduces steadily with more 'shots'. Otherwise the trends from LMx0 prevail. Bn and Hi are best partners, and both of them show similar performance as a source for Te. Incorporating all-but-target (ALL) languages gives further performance boost. The improvements show that **TransRel** can learn target-specific features with very few examples.

| ↓Source | Tasks | Bn | | | Hi | | | Te | | |
|---|---|---|---|---|---|---|---|---|---|---|
| | | LMx1 | LMx5 | LMx10 | LMx1 | LMx5 | LMx10 | LMx1 | LMx5 | LMx10 |
| En | RE | 68.89 | 78.37 | 82.30 | 72.25 | 78.95 | 82.48 | 47.12 | 57.11 | 57.40 |
| | ME | 73.30 | 81.97 | 84.03 | 75.72 | 83.82 | 84.72 | 51.04 | 58.91 | 63.20 |
| Bn | RE | – | – | – | 79.80 | 85.02 | 89.83 | 61.18 | 64.10 | 70.14 |
| | ME | – | – | – | 80.82 | 87.71 | 90.74 | 62.67 | 67.34 | 69.80 |
| Hi | RE | 77.64 | 83.38 | 84.55 | – | – | – | 64.24 | 63.98 | 68.21 |
| | ME | 79.73 | 84.51 | 84.02 | – | – | – | 64.02 | 68.51 | 71.72 |
| Te | RE | 71.78 | 78.48 | 83.73 | 73.34 | 78.22 | 83.76 | – | – | – |
| | ME | 73.54 | 79.52 | 82.23 | 71.57 | 79.83 | 85.21 | – | – | – |
| ALL | RE | 81.09 | 85.22 | 86.18 | 86.07 | 88.35 | 90.55 | 61.95 | 67.87 | 68.18 |
| | ME | 83.77 (-8.56) | 87.71 (-4.52) | 88.57 (-3.76) | 86.36 (-8.24) | 89.99 (-4.61) | 91.80 (-2.8) | 66.68 (-8.55) | 69.99 (-5.44) | 71.23 (-4.23) |
| LMx0 (best) | | | 81.06 | | | 83.54 | | | 62.00 | |

Table 6: Macro F1 of LMx{1,5,10} for different source and target languages. Green denotes performance gain (darker denotes larger gain) with respect to LMx0 (last row) for a given target language. Gap from ELFI for the best performing cell is in parentheses.

| ↓Target | Tasks | MTx0 | MTx1 | MTx5 | MTx10 |
|---|---|---|---|---|---|
| Bn | RE | 66.22 | 66.29 | 79.10 | 82.38 |
| | ME | 69.99 | 71.62 | 80.03 | 85.06 (-7.27) |
| Hi | RE | 73.35 | 77.62 | 84.05 | 86.46 |
| | ME | 74.84 | 81.75 | 85.95 | 87.14 (-7.46) |
| Te | RE | 51.89 | 51.13 | 58.87 | 61.34 |
| | ME | 50.25 | 58.81 | 60.75 | 66.37 (-9.06) |

Table 7: Macro F1 for MTx{0,1,5,10}. Performance gap between best performance (Green) and ELFI is shown in parentheses.

## 5.4 MTx performance

Here we investigate the importance of silver instances in a low-resource scenario. In the absence of abundant target instances for ELFI-style training, we can leverage potentially plentiful instances from a source language, namely, En. Table 7 shows the results.

- Even after using all available En gold instances to generate silver target instances, adding gold target instances gives steady improvements.
- At 10 instance per relation gold exposure, there is still a gap of 7.27% for Bn, 7.46% for Hi and 9.06% for Te compared to ELFI.

This gives a strong idea of the worth of gold target instances, and motivates *importing from the optimal languages* — an optimization that opens up a significant avenue of future work.

## 6 Performance of Ix

Here we compare the performance of instance transfer Ix with that of the model transfer LMx. We experiment using the zero-shot and 10-shot settings with RE and ME. In Table 8, for each setting, we report 10-shot performance with the gain obtained from its zero-shot counterpart in parentheses. Ix0 performs better than LMx0 for all source-target settings. High quality translations of well formed instances play a crucial role in the performance gain of instance transfer. However, the gain of 10-shot transfer over zero-shot in Ix is less than that of LMx. One possible reason could be that in the Ix setting, because we translate everything to English, few-shot examples hardly add any language specific information; while in the case of LMx, few-shot examples are likely to add more target language specific information to the model.

## 7 Augmented FewRel comparison

Here we investigate the effectiveness of extending an existing RE dataset for Indian languages. Existing supervised dataset **FewRel** (Han et al., 2018) has 27 common relations with **IndoRE**.

| ↓Target | FewRel | | IndoRE (ours) | |
|---|---|---|---|---|
| | Training instances | Macro F1 | Training instances | Macro F1 |
| | ELFI | | | |
| **Bn** | 18277 | 84.19 | 3845 | 92.46 |
| **Hi** | 18387 | 91.32 | 2987 | 96.45 |
| **Te** | 18462 | 74.66 | 5136 | 79.45 |
| | Ix | | | |
| **Bn** | 18900 | 80.26 | 3692 | 89.27 |
| **Hi** | 18900 | 90.78 | 2944 | 96.41 |
| **Te** | 18900 | 69.57 | 5101 | 80.38 |

Table 9: Macro F1 training with FewRel vs. IndoRE.

We translate the En evidence sentences released by them to Bn, Hi and Te using Google Translate API, and map the entities in the translated sentences. In Table 9 we report macro F1 obtained on the 27 relations while our model (**CLS+ES**) is trained with FewRel translated data, vs. our gold data in ELFI setting. We use 80% of our data to train and 20% to test. We use the whole FewRel data for 27 relations for training and use the same test split of ours to evaluate the performance. We observe that there is a performance gain greater than 5% in all the three languages, with highest gain achieved in Te, *while using far fewer labeled instances*. This underscores the necessity of developing RE datasets. In the instance transfer setup we translate the common relation instances of IndoRE to English and report the results in the last three rows of Table 9. We observe more than 6% gain in all languages.

| ↓Source | Task | Bn | | Hi | | Te | |
|---|---|---|---|---|---|---|---|
| | | LMx10 | Ix10 | LMx10 | Ix10 | LMx10 | Ix10 |
| En | RE | 82.30 (+ 18.93) | 84.14 (+ 9.60) | 82.48 (+ 14.83) | 86.30 (+ 7.39) | 57.40 (+ 14.87) | 70.18 (+ 9.36) |
| | ME | 84.03 (+ 17.19) | 84.29 (+ 9.06) | 84.72 (+ 15.35) | 87.05 (+ 6.10) | 63.20 (+ 18.17) | 71.01 (+ 9.73) |
| Bn | RE | - | - | 89.83 (+ 8.38) | 89.98 (+ 4.04) | 70.14 (+ 13.04) | 75.83 (+ 2.70) |
| | ME | - | - | 90.74 (+ 10.35) | 89.74 (+ 3.32) | 69.80 (+ 8.84) | 79.10 (+ 4.44) |
| Hi | RE | 84.55 (+ 8.32) | 85.61 (+ 4.78) | - | - | 68.21 (+ 11.43) | 75.36 (+ 0.26) |
| | ME | 84.02 (+ 8.93) | 85.95 (+ 4.65) | - | - | 71.72 (+ 16.58) | 78.73 (+ 2.83) |
| Te | RE | 83.73 (+ 23.45) | 81.94 (+ 6.32) | 83.76 (+ 23.28) | 86.54 (+ 7.83) | - | - |
| | ME | 82.23 (+ 20.11) | 83.53 (+ 10.98) | 85.21 (+ 18.84) | 87.19 (+ 6.89) | - | - |
| ALL | RE | 86.18 (+ 6.50) | 88.07 (+ 2.75) | 90.55 (+ 7.10) | 90.93 (- 0.62) | 68.18 (+ 6.60) | 78.74 (+ 2.10) |
| | ME | 88.57 (+ 7.51) | 89.08 (+ 1.98) | 91.80 (+ 8.26) | 91.66 (+ 1.02) | 71.23 (+ 9.23) | 80.37 (+ 1.26) |

Table 8: Macro F1 comparison between LMx10 and Ix10 for different source and target languages. The corresponding gains from zeroshot is reported in parentheses.

## 8 Transfer error analysis

We investigate different type of errors occurring in LMx and Ix setting. Table 10 shows some of the error types with examples. In case of LMx, we sometimes find that the prediction is sensitive to changes in one or two words, while the Ix counterpart is more robust, as we can see in the first example in Table 10. The second example shows that the words between the two entities also play a significant role in prediction; e.g., if a word strongly associated with the relation is present in between the two entities, the prediction is more accurate compared to when the word is at some other part of the sentence. Unavailability of diverse data while training the LMx model may be the main contributing factor for such errors. We build our RE model using mBERT, which has larger exposure to English compared to low-resource Indian languages. This possibly helps Ix perform better than LMx. Word reordering may help Google Translate to be more accurate, which in turn boosts Ix performance, as we see in the third example in Table 10. As we can see in the last example the major reason behind the occasional poor prediction of Ix is the incomplete or wrong translation with erroneous entity mapping from the target sentence.

## 9 Related work on RE systems

Named entity recognition and relation extraction have been of continued interest in recent years. In this section we discuss relevant work that propose different architectures for increasingly accurate named entity and relation prediction. Zheng et al. (2017) jointly extract entities and relations based on a novel tagging scheme. Chen et al. (2020) propose joint named entity recognition and relation extraction training in the clinical analytic space. In a similar direction, Wang and Lu (2020) propose table-sequence encoders for joint training. Bekoulis et al. (2018) use adversarial training for joint entity and relation extraction. Recently, a BERT-based model (Eberts and Ulges, 2019) used span-based encoding for joint extraction. All of these approaches show promising improvements in RE from joint NER+RE training in comparison to standalone RE for monolingual data. Fu et al. (2019) build a dependency graph and use a GCN to infer necessary information. The BiLSTM-CRF-based entity recognition model (Nguyen and Verspoor, 2019) with a deep biaffine attention layer to model second-order interactions among latent features for relation classification shows great performance for English relation extraction. However, these models incorporate specific building blocks (e.g., the dependency graph) which are not easy to procure with high accuracy in case of low-resource languages.

| Observation | Test Instance (Prediction - LMx) | Test Instance (Prediction - Ix) |
|---|---|---|
| Ix is robust to change of entities<br>Ground Truth: composer | আনন্দ সমারাকূন, (১৯১১–১৯৬২), শ্রীলঙ্কা কে এক সংগীতকার, কবি এবং শিক্ষক থে সমারাকূন নে শ্রীলঙ্কা কে রাষ্ট্রগান শ্রীলংকা মাতা কী রচনা কী থী ঔর উন্হে কলাত্মক শ্রীলংকাই সংগীত কা পিতা মানা জাতা হ্যায়<br>Prediction - screenwriter | Ananda Samarakoon, (1911–1962), was a Sri Lankan musician, poet and teacher. Samarakoon composed the Sri Lankan national anthem Sri Lanka Mata and is considered the father of artistic Sri Lankan music.<br>Prediction - composer |
| | আনন্দ সমারাকূন, (১৯১১–১৯৬২), শ্রীলঙ্কা কে এক সংগীতকার, কবি এবং শিক্ষক থে সমারাকূন নে শ্রীলঙ্কা কে রাষ্ট্রগান শ্রীলংকা পিতা কী রচনা কী থী ঔর উন্হে কলাত্মক শ্রীলংকাই সংগীত কা পিতা মানা জাতা হ্যায়<br>Prediction - child | Ananda Samarakoon,(1911–1962), was a Sri Lankan musician, poet and teacher. Samarakoon composed the Sri Lankan national anthem Sri Lanka Father and is considered the father of artistic Sri Lankan music.<br>Prediction composer |
| LMx prediction depends on word ordering<br>Ground Truth: child | হিন্দুদের ধর্মগ্রন্থ ও মহাকাব্য মহাভারত অনুযায়ী ভীষ্মক ছিলেন বিদর্ভের রাজা এবং কৃষ্ণের প্রথমা পত্নী রুক্মিণীর পিতা।<br>Prediction - mother | According to Hindu scriptures and the epic Mahabharata, Bhishmak was the king of Vidarbha and the father of Krishna's first wife Rukmini.<br>Prediction - child |
| | হিন্দুদের ধর্মগ্রন্থ ও মহাকাব্য মহাভারত অনুযায়ী বিদর্ভের রাজা এবং কৃষ্ণের প্রথমা পত্নী রুক্মিণীর পিতা ছিলেন ভীষ্মক<br>Prediction - child | According to the Hindu scriptures and the epic Mahabharata, Bhishmak was the father of Rukmini the king of Vidarbha and Krishna's first wife.<br>Prediction - child |
| Change in word ordering helps both LMx and Ix<br>Ground Truth: child | বলিউডে এখনও নতুন মুখ হিসাবেই বিবেচিত হন জাহ্নবী কাপুর মাত্র দুটি ছবি এবং একটি শর্টফিল্মে অভিনয় করেছেন শ্রীদেবী কন্যা<br>Prediction - spouse | Jahnavi Kapoor is still considered a new face in Bollywood. Sridevi Kanya has acted in only two films and one short film.<br>Prediction - performer |
| | বলিউডে এখনও নতুন মুখ হিসাবেই বিবেচিত হন শ্রীদেবী কন্যা জাহ্নবীকাপুর মাত্র দুটি ছবি এবং একটি শর্টফিল্মে অভিনয় করেছেন<br>Prediction - child | Sridevi's daughter Jahnavi Kapoor, who is still considered a new face in Bollywood, has acted in only two films and one short film.<br>Prediction - child |
| Ix prediction depends on translation quality<br>Ground Truth: award_received | কানাইলাল শীল (৫ সেপ্টেম্বর, ১৮৯৫-২০ জুলাই, ১৯৭৪) ছিলেন একজন বাঙালি দোতারাবাদক, সুরকার, লোকসঙ্গীত রচয়িতা ও সংগ্রাহক, লোকসঙ্গীতে অবদানের জন্য বাংলাদেশ সরকার তাকে ১৯৮৭ সালে মরণোত্তর একুশে পদক ভূষিত করে<br>Prediction - award_received | Kanailal Sheel (September 5, 1895 - July 20, 1984) was a Bengali dotara player, composer, composer and collector of folk music.<br>Prediction - occupation<br>Correct translation - Kanailal Sheel (September 5, 1895- July 20, 1974) was a Bengali dotaraplayer, composer, folk music composer and collector, who was posthumously awarded the Ekushey Medal by the Government of Bangladesh in 1987 for his contribution to folk music. |

Table 10: Error analysis of transfer mechanisms.

## 10 Conclusion

In this work, we release **IndoRE**, a rich set of gold-tagged sentences for NER and relation extraction in En, Bn, Hi and Te. We also present **TransRel**, a multilingual joint NER and RE system with interlingual transfer. Our experiments show promising results on generalization and knowledge transfer. Even in the absence of gold training data for a given language, instances from other languages can contribute to considerable RE accuracy boosts. We uncover intuitive patterns in the extent of improvements conferred by one language on another. We believe IndoRE and TransRel can significantly assist future study of Indian language relation extraction.

# A Data Bootstrapping Recipe for Low-Resource Multilingual Relation Classification
## (Appendix)

## A Complexity of relations

Figure 3 shows the average lexical distance between entities accross relations as a measure of complexity.

## B Annotating translated sentences

Starting from gold annotated sentences in the source language, it is nontrivial to transfer or 'project' the gold entity spans to the sentence after translation to the target language. The steps we follow to do this are shown in Algorithm 1.

**Algorithm 1** Re-annotate translated sentence with entity spans.
---
1: **Input:** Translated sentence, entity1, entity2 as $S^{tr}$, $E_1^{tr}$, $E_2^{tr}$ respectively.
2: **Output:** Entity annotated translated sentence $S_a^{tr}$
3: find the word count for $S^{tr}$, denote it as $n$.
4: **for** each $E_i^{tr}$, $\forall i \in \{1,2\}$ **do**
5:    Find the number of words in it, denote it as $l$.
6:    Divide the sentence of $n$ word count into $n - l + 1$ contiguous overlapping windows.
7:    Find the window that has minimum Levenshtein distance with $E_i^{tr}$, denote it as $w_i$.
8:    Add $[E_i]$ as prefix to the first word in $w_i$, and add $[/E_i]$ as suffix to the last word in $w_i$.
9: **end for**
10: return entity annotated translated sentence $S_a^{tr}$.
---

## C Base model variations

The best performing configuration of the base RE system is described in Figure 2. In this section, we briefly describe the other variations we evaluated.

### C.1 Input layer

We append special markers to the input sentence ($s$) for useful embedding generation by mLMs in different ways other than the best performing one reported in main paper.

**Entity type markers (ET):** Mark entity spans by special tokens denoting their type as follows:
$$s = ([CLS], w_1, \ldots, [ET_1], entity, [/ET_1],$$
$$w_i, \ldots, [ET_2], entity, [/ET_2], \ldots, [SEP])$$

Here $ET_1, ET_2$ are the entity types of the corresponding entities.

**Entity span and type markers (EST):** We mark both the entity number (1 or 2) and their types:
$$s = ([CLS], w_1, \ldots, [E_1], [ET_1], entity,$$
$$[/ET_1], [/E_1], w_i, \ldots, [E_2], [ET_2],$$
$$entity, [/ET_2], [/E_2], \ldots, [SEP])$$

**Language identity marker (L):** We append the language identifier token $[L]$, where $L$ is in {En,Bn,Hi,Te}) after the [CLS] token for all of the above cases:
$$s = ([CLS], [L], w_1, \ldots, [E_1], entity,$$
$$[/E_1], w_i, \ldots, [E_2], entity, [/E_2],$$
$$\ldots, [SEP])$$

### C.2 Embedding layer

We experimented with three different multilingual language models (mLMs), namely, mBERT[9], XLM-R (Conneau et al., 2020) and IndicBERT (Kakwani et al., 2020) to leverage the pretrained multilingual embedding space. mBERT was generally the best.

### C.3 Summarization layer

The mLM generates embeddings ($h$) for each word as well as special tokens introduced in different input formats given a sentence $s$. As these mLMs provide contextual embeddings, they are expected to emit useful and different information corresponding to different special tokens. Here we describe several information pooling mechanisms to evaluate the effectiveness of special tokens, other than the ones described in the main paper.

**CLS:** Here we consider CLS token embedding which encodes necessary contextual information about the sentence.
$$h_s = h_{[CLS]}$$

**ES:** Here we only consider the entity specific information from special entity span tokens $[E_1 \ldots /E_1]$ and $[E_2 \ldots /E_2]$.
$$h_s = [h_{E_1}, h_{E_2}]$$

## D Other performance measures

In the main paper we reported macro averaged F1 across labels. Here we also provide micro averaged 0/1 accuracy metric. The broad trends remain the same. In the Table 11 we report the accuracy for ELFI setting in three languages. In the Table 12 we report the accuracy obtain in different source-target language pair in zero-shot setting. Table 15 shows the accuracy for LMx few-shot transfer. Table 16 shows the accuracy comparison between LMx0,10 with Ix0,10. Table 13 shows zero-shot, few-shot accuracy for different target languages in MTx setting. And Table 14 shows the accuracy comparison between FewRel and IndoRE for both ELFI and Ix.

---
[9] https://github.com/google-research/bert/blob/master/multilingual.md

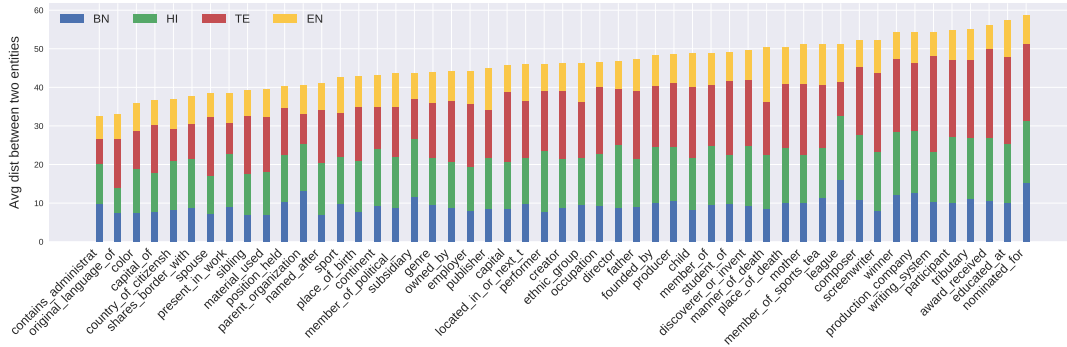

Figure 3: Average lexical distance between entities across relation and languages. All relations in En and Bn have small lexical distance. Most relations need large lexical distances for Te and Hi.

| ↓Target | FewRel | | IndoRE (ours) | |
|---|---|---|---|---|
| | Training instances | Accuracy | Training instances | Accuracy |
| ELFI | | | | |
| Bn | 18277 | 85.20 | 3845 | 92.39 |
| Hi | 18387 | 91.82 | 2987 | 96.46 |
| Te | 18462 | 76.81 | 5136 | 85.55 |
| Ix | | | | |
| Bn | 18900 | 82.93 | 3692 | 89.93 |
| Hi | 18900 | 91.86 | 2944 | 96.55 |
| Te | 18900 | 74.60 | 5101 | 85.71 |

Table 14: Accuracy training with FewRel vs. IndoRE.

| Summerization Layer | Model | BN | HI | TE |
|---|---|---|---|---|
| CLS+ES | RE | 93.01 | 95.13 | 85.91 |
| | ME | 93.75 | 95.13 | 87.12 |
| CLS+EN | RE | 92.83 | 94.69 | 85.91 |
| | ME | 93.21 | 95.42 | 86.94 |

Table 11: Microaveraged 0/1 accuracy of ELFI.

| Train | Model | BN | HI | TE |
|---|---|---|---|---|
| self | | 93.75 | 95.42 | 87.12 |
| EN | RE | 65.75 | 71.82 | 52.02 |
| | ME | 70.96 | 75.38 | 55.87 |
| BN | RE | - | 85.40 | 67.85 |
| | ME | - | 84.10 | 68.09 |
| HI | RE | 77.70 | - | 61.59 |
| | ME | 78.00 | - | 62.01 |
| TE | RE | 64.64 | 67.03 | - |
| | ME | 66.48 | 72.69 | - |
| ALL | RE | 81.19 | 87.22 | 68.33 |
| | ME | 82.48 | 86.93 | 70.62 |

Table 12: Performance (microaveraged accuracy) of zero-shot training on different languages. Green represents closest performance and Red represents the most deviating performance compared to self training. Yellow represents the best performance among single language input.

| ↓Target | Task | 0-shot | 1 shot | 5 shot | 10 shot |
|---|---|---|---|---|---|
| BN | RE | 70.04 | 69.18 | 80.70 | 84.01 |
| | ME | 72.79 | 74.75 | 81.00 | 86.34 (-7.41) |
| HI | RE | 76.32 | 80.75 | 85.62 | 87.51 |
| | ME | 78.50 | 84.46 | 88.02 | 88.67 (-6.75) |
| TE | RE | 60.26 | 60.63 | 69.96 | 69.90 |
| | ME | 57.62 | 66.83 | 69.90 | 73.45 (-13.67) |

Table 13: Performance (microaveraged accuracy) for silver training (MTx) for different target languages. The gaps between the best performance and ELFI are shown in parentheses.

| ↓ Source | Tasks | Bn | | | Hi | | | Te | | |
|---|---|---|---|---|---|---|---|---|---|---|
| | | LMx1 | LMx5 | LMx10 | LMx1 | LMx5 | LMx10 | LMx1 | LMx5 | LMx10 |
| En | RE | 71.57 | 80.02 | 83.95 | 76.69 | 81.84 | 84.46 | 55.57 | 61.29 | 64.90 |
| | ME | 75.12 | 83.88 | 84.99 | 80.10 | 85.19 | 86.20 | 61.53 | 65.86 | 70.74 |
| Bn | RE | – | – | – | 83.88 | 86.71 | 90.85 | 67.43 | 69.48 | 75.68 |
| | ME | – | – | – | 84.82 | 88.67 | 91.79 | 69.72 | 70.80 | 76.94 |
| Hi | RE | 79.66 | 85.23 | 85.97 | – | – | – | 68.93 | 69.66 | 72.43 |
| | ME | 82.05 | 85.42 | 85.48 | – | – | – | 68.99 | 71.88 | 75.14 |
| Te | RE | 73.65 | 80.15 | 85.11 | 78.29 | 81.70 | 86.27 | – | – | – |
| | ME | 74.82 | 80.76 | 83.64 | 77.05 | 82.43 | 87.36 | – | – | – |
| ALL | RE | 82.10 | 86.64 | 87.68 | 88.23 | 89.54 | 91.50 | 70.38 | 73.15 | 76.10 |
| | ME | 84.74 | 88.42 | 89.83 | 87.73 | 91.19 | 92.74 | 72.07 | 74.17 | 78.75 |
| | | (-9.01) | (-5.33) | (-3.92) | (-7.69) | (-4.23) | (-2.68) | (-15.05) | (-12.95) | (-8.37) |
| LMx0 (best) | | | 82.48 | | | 87.22 | | | 70.62 | |

Table 15: Accuracy of LMx{1,5,10} for different source and target languages. Green denotes performance gain (darker denotes larger gain) with respect to LMx0 (last row) for a given target language. Gap from ELFI for the best performing cell is in parentheses.

| ↓Source | Task | Bn | | Hi | | Te | |
|---|---|---|---|---|---|---|---|
| | | LMx10 | Ix10 | LMx10 | Ix10 | LMx10 | Ix10 |
| **En** | RE | 83.95 (+ 18.25) | 85.43 (+ 8.99) | 84.46 (+ 12.64) | 88.69 (+ 6.28) | 64.90 (+ 12.88) | 77.08 (+ 9.22) |
| | ME | 84.99 (+ 14.03) | 85.50 (+ 8.62) | 86.20 (+ 10.82) | 88.69 (+ 5.76) | 70.74 (+ 14.87) | 77.47 (+ 8.13) |
| **Bn** | RE | - | - | 90.85 (+ 5.45) | 90.91 (+ 2.59) | 75.68 (+ 7.83) | 80.89 (+ 4.27) |
| | ME | - | - | 91.79 (+ 7.69) | 90.47 (+ 2.00) | 76.94 (+ 8.85) | 81.85 (+ 1.99) |
| **Hi** | RE | 85.97 (+ 8.21) | 87.27 (+ 3.69) | - | - | 72.43 (+ 10.84) | 79.52 (+ 0.62) |
| | ME | 85.48 (+ 7.48) | 87.21 (+ 3.04) | - | - | 75.14 (+ 13.13) | 81.00 (+ 1.36) |
| **Te** | RE | 85.11 (+ 20.47) | 83.91 (+ 6.08) | 86.27 (+ 19.24) | 88.40 (+ 5.47) | - | - |
| | ME | 83.64 (+ 17.16) | 84.93 (+ 7.98) | 87.36 (+ 14.67) | 88.91 (+ 5.54) | - | - |
| ALL | RE | 87.68 (+ 6.49) | 89.30 (+ 2.41) | 91.50 (+ 4.28) | 92.61 (+ 0.15) | 76.10 (+ 7.77) | 82.48 (+ 1.76) |
| | ME | 89.83 (+ 7.43) | 89.99 (+ 1.50) | 92.74 (+ 5.81) | 92.90 (+ 0.98) | 78.75 (+ 8.13) | 83.96 (+ 1.99) |

Table 16: Accuracy comparison between LMx10 and Ix10 for different source and target languages. The corresponding gains from zeroshot is reported in parentheses.